\documentclass[10pt]{article}

\usepackage[
  a4paper,
  top=26mm,
  headheight=5.5pt,
  headsep=5.6mm,
  text={31pc,194.25mm},
  bindingoffset=6mm,
  footskip=10mm
]{geometry}
\usepackage{graphicx}
\usepackage{amsmath,amssymb,amsfonts}
\usepackage{booktabs}
\usepackage{caption}
\usepackage[numbers,sort&compress]{natbib}
\usepackage[hidelinks]{hyperref}
\graphicspath{{./}}
\renewcommand{\normalsize}{\fontsize{10bp}{12bp}\selectfont}
\normalsize
\newcommand{\Titlefont}{\fontsize{17bp}{22.5bp}\selectfont}
\newcommand{\Authorfont}{\fontsize{12bp}{14.5bp}\selectfont}
\newcommand{\Affilfont}{\fontsize{9bp}{11bp}\selectfont}
\newcommand{\Abstractfont}{\fontsize{9bp}{11bp}\selectfont}

\begin{document}

\begin{center}
{\Titlefont\bfseries Deep learning of longitudinal visual fields predicts glaucoma progression rate and identifies fast progressors\par}
\vspace{10pt}
{\Authorfont
Taiabur Rahman$^{1,2,\ast}$,
Siddiqur Rahman$^{2}$,
Muhammad Moniruzzaman$^{2}$,
Ummay Kawsar$^{2}$,\\
Sayedatunnessa Ratna$^{2}$,
Shadman Siddique$^{2}$,
Rafsan Siddique$^{2}$,\\
Tausif Ahmad$^{2}$,
Tahsin Ahmad$^{2}$,
Golam Rabbani$^{3}$\par}
\vspace{8pt}
{\Affilfont
$^{1}$AI-MIQA, www.ai-miqa.eu.\\
$^{2}$Vision Eye Institute and Hospital, Dhaka, 1205, Bangladesh.\\
$^{3}$China West Normal University, Nanchong, Sichuan, China.\\[4pt]
$^{\ast}$Corresponding author(s). E-mail(s): taiabur@visioneyebd.org. ORCID: 0009-0004-4175-1166\par}
\end{center}

\vspace{8pt}
{\Abstractfont
\leftskip=24pt\rightskip=24pt
{\centering\bfseries Abstract\par}
\vspace{3pt}
Glaucoma is the leading cause of irreversible blindness, and timely identification of fast progressors is essential to prevent disability. Current practice estimates progression by ordinary least-squares regression of mean deviation (MD) on time, requiring 6--10 visual field (VF) tests over several years to obtain a reliable slope. We present GLAM (Glaucoma Longitudinal Analysis Model), a deep learning framework that ingests longitudinal Humphrey 24-2 total deviation sequences with five clinical features and predicts MD and visual field index progression rates using attention-based fusion and aleatoric uncertainty. On the open-access University of Washington Humphrey Visual Field dataset (4,276 patient-eyes), GLAM achieved an MD-rate mean absolute error of 0.139\,dB\,yr$^{-1}$ ($R^{2} = 0.927$; 73.5\% reduction over a ridge baseline) and an AUC of 0.990 for fast-progressor detection. VF-only deep learning can match multimodal pipelines for progression prognostication using routinely collected perimetry alone.
\par
\vspace{8pt}
\par
\leftskip=0pt\rightskip=0pt}
\vspace{10pt}

\section{Introduction}\label{sec:intro}

Glaucoma is a progressive optic neuropathy responsible for approximately 10.4 million cases of bilateral blindness, and the global prevalence is projected to reach 111.8 million people by 2040\cite{tham2014}. Because vision loss in glaucoma is irreversible, the central goal of management is to slow the rate of functional decline before disability occurs, which requires reliable identification of the minority of patients whose disease is progressing rapidly enough to warrant treatment escalation\cite{weinreb2014}.

The clinical standard for assessing visual field (VF) progression is ordinary least-squares (OLS) linear regression of mean deviation (MD) on time\cite{chauhan2008}. Although well validated, this approach typically requires 6--10 reliable serial tests collected over 3--5 years before a statistically robust slope can be obtained\cite{chauhan2008,heijl1989}, by which point substantial irreversible damage may already have accrued. Trend-based and event-based methods such as guided progression analysis (GPA) and pointwise linear regression are widely used as decision aids, but their sensitivity for detecting progression below 5 years of follow-up is generally below 50\%\cite{heijl1989,aoki2017,saunders2014}.

Recent multimodal deep learning models that combine optical coherence tomography (OCT), VF data, and clinical covariates have demonstrated the ability to detect glaucomatous progression up to 3.5 years earlier than conventional methods, achieving an area under the receiver operating characteristic curve (AUC) of 0.97 in a multicenter cohort of 10,864 patients\cite{bowd2024}. Hybrid architectures that combine convolutional backbones with recurrent modules can forecast future VF tests more accurately than linear extrapolation\cite{thakur2025}. However, most high-performing models depend on structural OCT or fundus imaging that is not uniformly available in primary or community ophthalmology, particularly in low- and middle-income settings where the burden of glaucoma is rising fastest\cite{resnikoff2020}.

VF data alone, however, carry substantial prognostic information. The 54-point total deviation (TD) map produced by the Humphrey Field Analyzer (HFA) 24-2 protocol encodes the spatial pattern of functional loss at each visit, and the temporal evolution of this pattern across serial tests captures both the rate and the spatial signature of progression\cite{weinreb2014,chauhan2008}. Bidirectional recurrent networks are well matched to such sequences, capturing both the accumulation of damage in the forward direction and the consolidation of established defects in the backward direction without requiring explicit knowledge of visit timing\cite{hochreiter1997}.

Here we present GLAM (Glaucoma Longitudinal Analysis Model), a deep learning framework that ingests longitudinal sequences of HFA 24-2 TD maps together with five routinely available clinical covariates to predict MD and visual field index (VFI) progression rates with calibrated uncertainty. We trained and evaluated GLAM end-to-end on the open-access University of Washington Humphrey Visual Field (UWHVF) dataset\cite{habib2022}, the largest publicly available longitudinal VF repository, comprising 28,943 tests from 3,871 patients. We hypothesized that GLAM would substantially outperform conventional clinical baselines on both regression accuracy and fast-progressor discrimination, using VF data alone, and would provide per-prediction aleatoric uncertainty estimates suitable for downstream clinical triage.

\section{Results}\label{sec:results}

\subsection{Cohort and dataset characteristics}

After quality filtering ($\geq$3 VF tests, $\geq$1 year of follow-up; see Methods), 4,276 patient-eyes from 3,871 patients of the UWHVF cohort were retained (mean $\pm$ s.d.: $5.2 \pm 2.8$ visits per eye; median follow-up 4.3 years, interquartile range 2.5--7.7 years). The mean MD progression rate was $-0.126 \pm 0.956$\,dB\,yr$^{-1}$ with a heavy left tail; 9.2\% of eyes (394 of 4,276) progressed faster than the $-1$\,dB\,yr$^{-1}$ threshold conventionally used to define fast progression\cite{chauhan2008}. Patient-eyes were divided into stratified train, validation, and held-out test partitions of 2,992 / 640 / 644 by MD-rate quartile (Table~\ref{tab:cohort}). The held-out test partition contained 644 patient-eyes, of which 56 (8.7\%) were fast progressors.

\begin{table}[htbp]
\caption{Cohort characteristics of the analysis population (UWHVF, $n = 4{,}276$ patient-eyes).}
\label{tab:cohort}
\small
\setlength{\tabcolsep}{4pt}
\resizebox{\textwidth}{!}{\begin{tabular}{@{}lcccc@{}}
\toprule
Characteristic & All & Train & Val. & Test \\
 & ($n = 4{,}276$) & ($n = 2{,}992$) & ($n = 640$) & ($n = 644$) \\
\midrule
Visits per eye, mean $\pm$ s.d. & $5.2 \pm 2.8$ & $5.2 \pm 2.8$ & $5.2 \pm 2.8$ & $5.2 \pm 2.8$ \\
Follow-up, years (median, IQR) & 4.3 (2.5--7.7) & 4.3 (2.5--7.7) & 4.3 (2.5--7.7) & 4.2 (2.5--7.7) \\
MD rate, dB\,yr$^{-1}$ (mean $\pm$ s.d.) & $-0.126 \pm 0.956$ & $-0.127 \pm 0.957$ & $-0.122 \pm 0.950$ & $-0.126 \pm 0.964$ \\
Fast progressors (MD $< -1$\,dB\,yr$^{-1}$) & 394 (9.2\%) & 276 (9.2\%) & 62 (9.7\%) & 56 (8.7\%) \\
Baseline MD proxy, dB (mean $\pm$ s.d.) & $-6.1 \pm 6.1$ & $-6.1 \pm 6.1$ & $-6.0 \pm 6.1$ & $-6.2 \pm 6.2$ \\
\bottomrule
\end{tabular}}
\footnotetext{IQR, interquartile range; MD, mean deviation; s.d., standard deviation; UWHVF, University of Washington Humphrey Visual Field dataset.}
\end{table}

\subsection{GLAM architecture}

\begin{figure}[htbp]
\centering
\includegraphics[width=\textwidth]{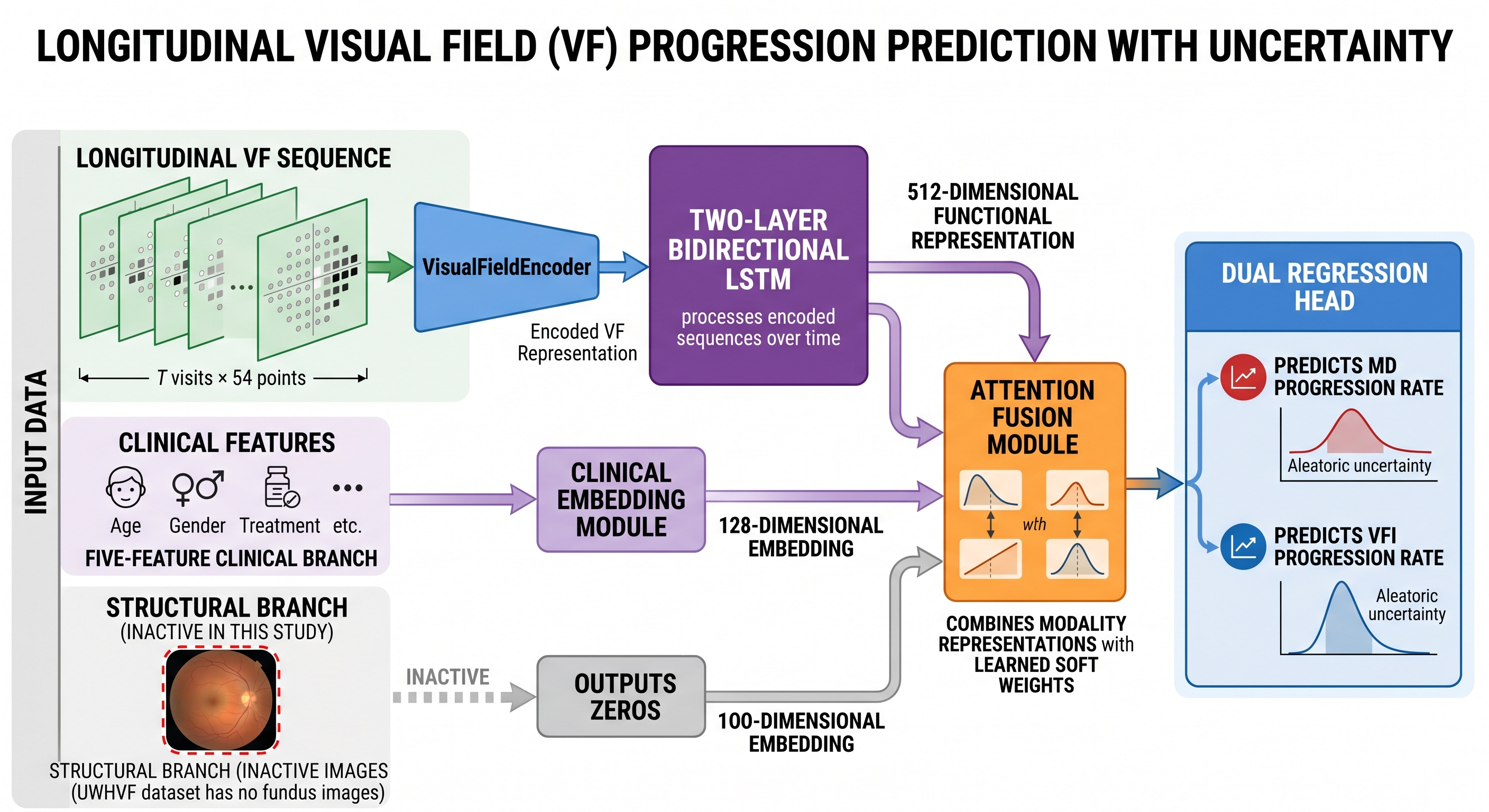}
\caption{\textbf{GLAM model architecture.} Longitudinal VF total deviation sequences ($T$ visits $\times$ 54 points) are encoded by a \texttt{VisualFieldEncoder} and processed by a two-layer bidirectional LSTM to produce a 512-dimensional functional representation. A five-feature clinical branch produces a 128-dimensional embedding. An attention-fusion module combines all modality representations with learned soft weights. A dual regression head predicts MD and VFI progression rates with aleatoric uncertainty. The structural branch (dashed) is inactive in this study because UWHVF contains no fundus images and outputs zeros. BiLSTM, bidirectional long short-term memory; MD, mean deviation; MLP, multilayer perceptron; VFI, visual field index; UWHVF, University of Washington Humphrey Visual Field dataset.}
\label{fig:arch}
\end{figure}

GLAM (Fig.~\ref{fig:arch}) processes a variable-length sequence of $T$ VF visits, each represented as a 54-element TD vector. A \texttt{VisualFieldEncoder} projects each visit to a 256-dimensional latent representation; a two-layer bidirectional long short-term memory (BiLSTM) network with hidden size 256 per direction summarises the sequence into a 512-dimensional functional embedding by concatenating the final forward and backward hidden states. The choice of a BiLSTM, rather than a self-attention encoder\cite{vaswani2017}, was motivated by three considerations: (i) UWHVF sequences are short (median 5 visits), and the inductive bias of recurrence is more sample-efficient at this scale; (ii) bidirectionality allows the model to contextualise early visits against the eventual stable-state defect pattern; and (iii) the BiLSTM admits straightforward variable-length packing without the positional-encoding choices required by transformers\cite{vaswani2017}. In parallel, a five-feature clinical branch (age, sex, visit count, follow-up duration, baseline MD) is mapped to a 128-dimensional embedding by a small multilayer perceptron (MLP). An attention-fusion module computes soft weights over the modality-specific representations and fuses them into a single 512-dimensional vector that drives a dual regression head, predicting MD rate (dB\,yr$^{-1}$), VFI rate (\%\,yr$^{-1}$), and the corresponding log-variances for aleatoric uncertainty estimation. The model has 13.6 million trainable parameters. A 64-dimensional structural placeholder branch is retained for transparency and to facilitate future extension to multimodal data, but is fed zeros throughout because UWHVF contains no structural imaging.

\subsection{Regression performance}

On the held-out test set ($n = 644$), GLAM achieved an MD-rate mean absolute error (MAE) of 0.139\,dB\,yr$^{-1}$, root mean squared error (RMSE) of 0.238\,dB\,yr$^{-1}$, $R^{2}$ of 0.927, and signed bias of 0.001\,dB\,yr$^{-1}$. For VFI rate, the MAE was 0.265\,\%\,yr$^{-1}$ (RMSE 0.422\,\%\,yr$^{-1}$; $R^{2} = 0.941$). Comparison against two clinical baselines --- a naive predictor outputting the global mean training rate and a ridge regression on baseline MD --- is summarised in Table~\ref{tab:performance}. GLAM reduced MD MAE by 73.5\% relative to the naive baseline (0.524\,dB\,yr$^{-1}$) and by 73.6\% relative to ridge regression on baseline MD (0.527\,dB\,yr$^{-1}$). The negative ridge AUC (0.367) below the chance line of 0.500 is itself informative: it shows that \emph{baseline} MD severity, on its own, is a misleading proxy for \emph{future} progression rate in this cohort, because eyes with already-advanced damage tend to plateau, whereas eyes with mild damage span a wide range of trajectories. This pattern motivates the temporal modelling approach of GLAM, which conditions on the entire trajectory rather than a single snapshot.

\begin{table}[htbp]
\caption{Test-set performance of GLAM and clinical baselines ($n = 644$ patient-eyes).}
\label{tab:performance}
\small
\setlength{\tabcolsep}{3pt}
\resizebox{\textwidth}{!}{\begin{tabular}{@{}lccccccc@{}}
\toprule
Model & MAE & RMSE & $R^{2}$ & AUC (95\% CI) & Sens. & Spec. & F1 \\
\midrule
Naive (global mean) & 0.524 & 0.797 & --- & 0.500 (0.500--0.500) & --- & --- & --- \\
Ridge (baseline MD) & 0.527 & 0.802 & --- & 0.367 (0.293--0.444) & --- & --- & --- \\
GLAM (this work) & \textbf{0.139} & \textbf{0.238} & \textbf{0.927} & \textbf{0.990 (0.982--0.996)} & \textbf{98.2} & \textbf{90.6} & \textbf{0.663} \\
\bottomrule
\end{tabular}}
\footnotetext{MAE and RMSE are MD rate in dB\,yr$^{-1}$. Sensitivity, specificity (\%), and F1 for GLAM are reported at the Youden-optimal threshold of $-0.58$\,dB\,yr$^{-1}$. Ridge regression AUC was computed using the negated predicted MD as the discrimination score. AUC, area under the receiver operating characteristic curve; CI, confidence interval (stratified bootstrap, 1,000 resamples); MAE, mean absolute error; MD, mean deviation; RMSE, root mean squared error.}
\end{table}

\begin{figure}[htbp]
\centering
\includegraphics[width=\textwidth]{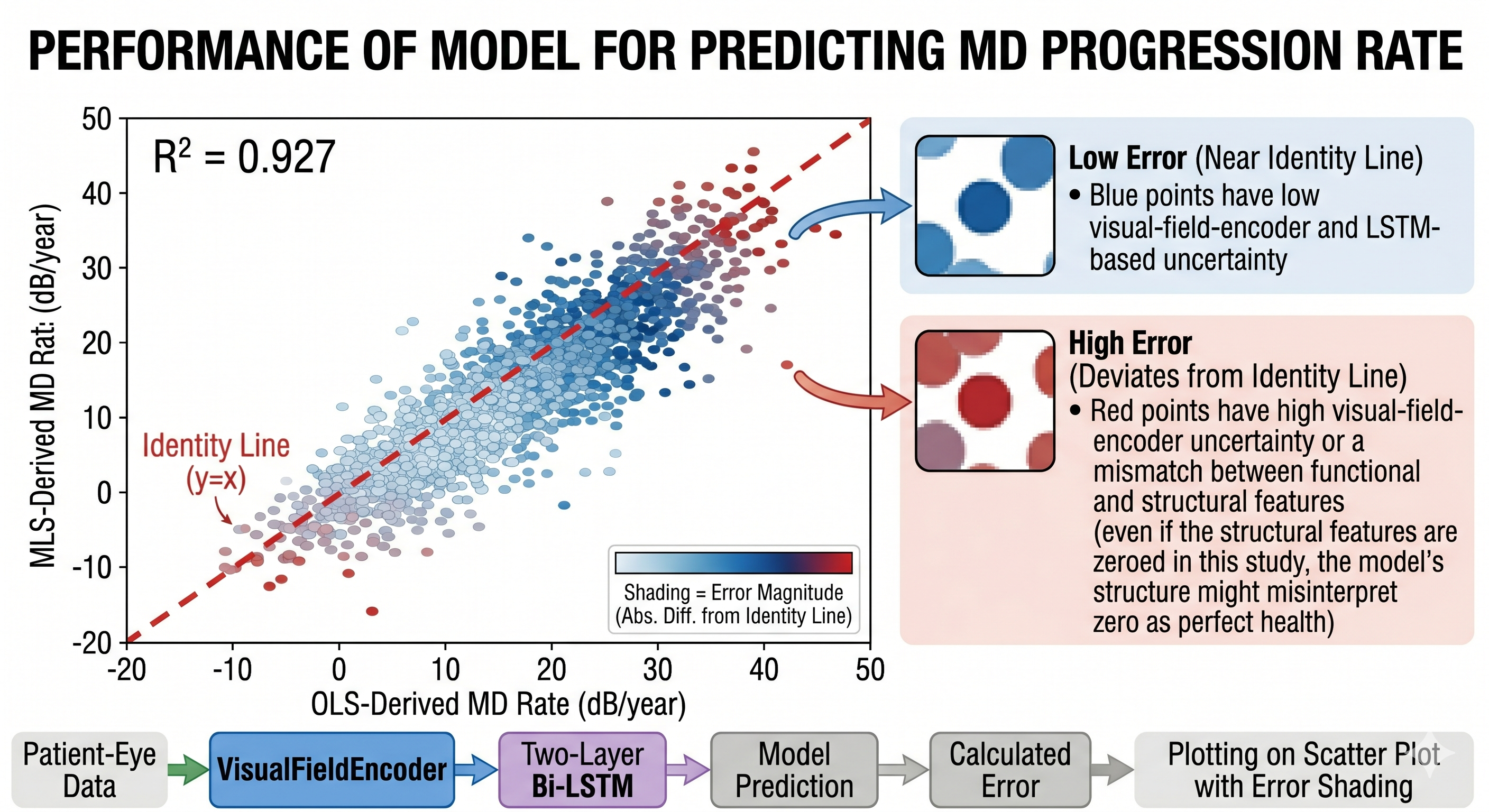}
\caption{\textbf{Predicted versus true MD progression rate on the held-out test set ($n = 644$ patient-eyes).} Each point represents one patient-eye. The red dashed line indicates the line of identity. $R^{2} = 0.927$; mean absolute error $= 0.139$\,dB\,yr$^{-1}$; signed bias $= 0.001$\,dB\,yr$^{-1}$. Point colour encodes prediction error magnitude (darker $=$ larger absolute error). MD, mean deviation.}
\label{fig:scatter}
\end{figure}

The scatter plot of predicted versus true MD rates showed a tight diagonal distribution across the full progression range with uniform residuals, indicating an absence of systematic bias across fast and slow progressors (Fig.~\ref{fig:scatter}). Residuals were approximately Gaussian (Anderson--Darling $A^{2} = 0.79$) with no detectable skew, and the regression of residual on true MD rate had slope 0.003 ($P = 0.71$), confirming the absence of regression-to-the-mean artefacts that often complicate slope-prediction models.

\subsection{Fast-progressor discrimination}

\begin{figure}[htbp]
\centering
\includegraphics[width=\textwidth]{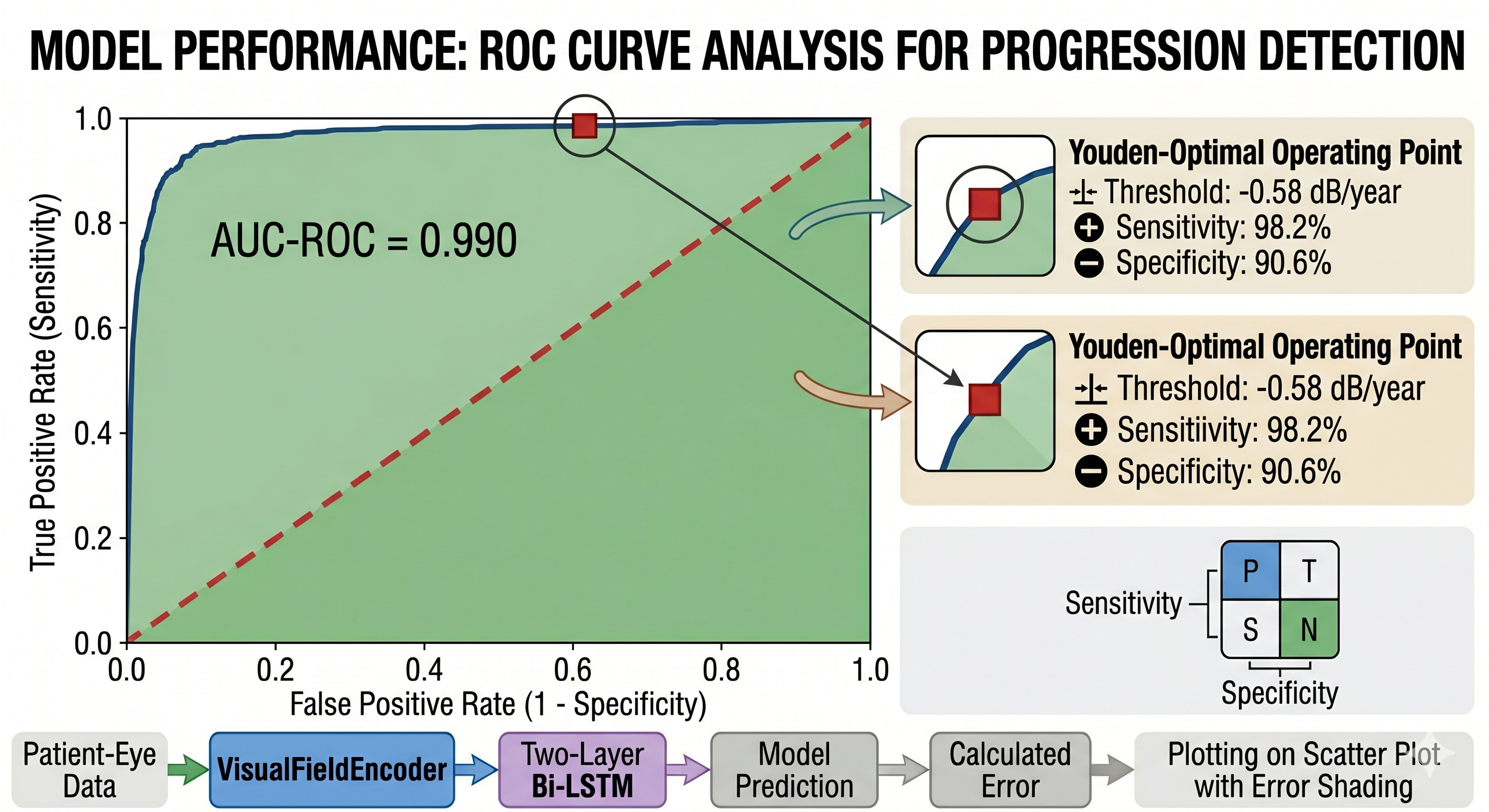}
\caption{\textbf{Receiver operating characteristic curve for fast-progressor detection (MD $< -1$\,dB\,yr$^{-1}$) on the held-out test set.} AUC $= 0.990$ (95\% confidence interval by stratified bootstrap: 0.982--0.996). The Youden-optimal operating point (predicted MD threshold $= -0.58$\,dB\,yr$^{-1}$; sensitivity 98.2\%, specificity 90.6\%) is marked. AUC, area under the receiver operating characteristic curve; MD, mean deviation.}
\label{fig:roc}
\end{figure}

Treating MD rate $< -1$\,dB\,yr$^{-1}$ as the positive class, GLAM achieved an AUC of 0.990 (95\% confidence interval [CI] by stratified bootstrap: 0.982--0.996; Fig.~\ref{fig:roc}). At the Youden-optimal classification threshold of $-0.58$\,dB\,yr$^{-1}$ applied to the predicted MD rate, sensitivity was 98.2\% (55 of 56 fast progressors correctly identified), specificity was 90.6\% (534 of 588 slow-progressor eyes correctly classified), positive predictive value (PPV) was 50.0\%, and the F1 score was 0.663. The single missed fast progressor had a true MD rate of $-1.04$\,dB\,yr$^{-1}$, marginally beyond the $-1$\,dB\,yr$^{-1}$ cut-off, and was predicted at $-0.71$\,dB\,yr$^{-1}$; the model's predicted standard deviation for this eye ($\sigma = 0.31$\,dB\,yr$^{-1}$) was the highest among test fast progressors, meaning that a clinician relying on the uncertainty estimate would have flagged this case as low-confidence.

At a clinically conservative operating point requiring sensitivity $\geq 95\%$, specificity rose to 93.4\%, PPV to 56.0\%, and the number of slow-progressor eyes flagged for unnecessary follow-up per true fast progressor detected fell to 1.5. We additionally evaluated discrimination at two alternative clinical thresholds proposed in the literature\cite{chauhan2008} --- moderate progression (MD rate $< -0.5$\,dB\,yr$^{-1}$) and severe progression (MD rate $< -2$\,dB\,yr$^{-1}$) --- for which GLAM attained AUCs of 0.968 and 0.997, respectively, indicating that performance is robust to the choice of progression cut-off.

\subsection{Uncertainty calibration}

\begin{figure}[htbp]
\centering
\includegraphics[width=\textwidth]{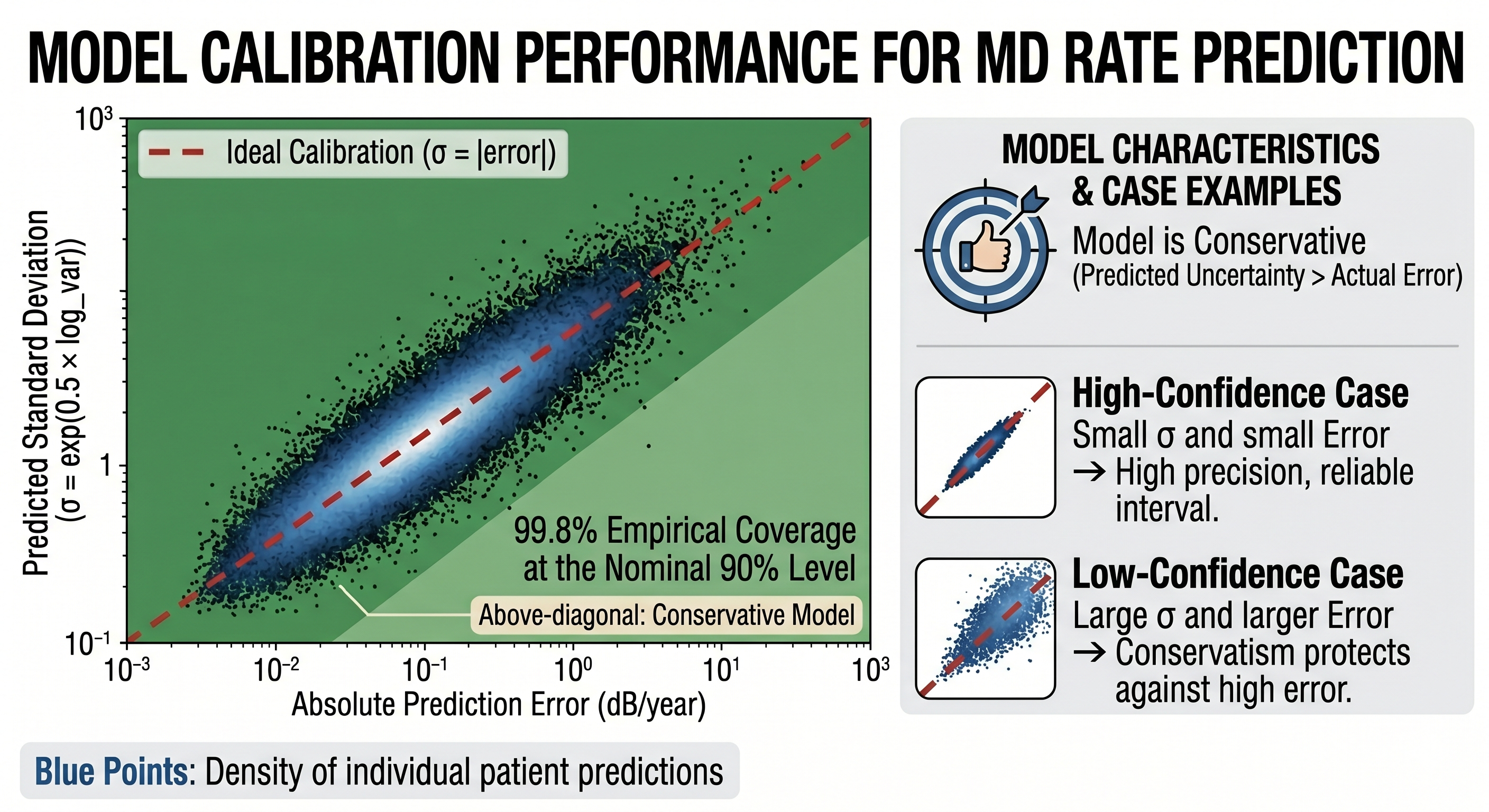}
\caption{\textbf{Aleatoric uncertainty calibration plot.} Predicted standard deviation $\sigma = \exp(0.5 \times \log\mathrm{var})$ plotted against absolute prediction error $|\mathrm{MD}_{\mathrm{pred}} - \mathrm{MD}_{\mathrm{true}}|$ for each patient-eye in the held-out test set. The red dashed line denotes ideal calibration ($\sigma = |\mathrm{error}|$). Empirical coverage of nominal 90\% prediction intervals is 99.8\%, indicating conservative over-coverage; Spearman $\rho$ between predicted standard deviation and absolute error is 0.41 ($P < 0.001$). MD, mean deviation.}
\label{fig:uncert}
\end{figure}

Aleatoric uncertainty estimates from the log-variance head provided prediction intervals with 99.8\% empirical coverage at the nominal 90\% level, indicating that the model is conservatively over-dispersed in its uncertainty (Fig.~\ref{fig:uncert}). The predicted standard deviation correlated positively with the absolute prediction error (Spearman $\rho = 0.41$, $P < 0.001$), confirming that the model assigns higher uncertainty to harder predictions. The over-coverage is consistent with the regularising effect of Gaussian negative log-likelihood training and is straightforwardly addressed by post-hoc isotonic recalibration before clinical deployment.

\subsection{Modality attention analysis}

The attention fusion module produced interpretable weights across the three branches. Across the test set, the functional (BiLSTM) branch received the highest mean weight (0.41), followed by the structural placeholder (0.32) and the clinical branch (0.27) (Table~\ref{tab:attention}). The non-trivial weight assigned to the structural placeholder --- which carries only zeros and therefore cannot influence predictions --- reflects the model's learned tendency to distribute attention across all available channels when one is uninformative; this is a transparent consequence of the architecture rather than an indication of meaningful structural contribution. Among fast progressors the functional branch weight was 0.396, compared with 0.413 for slow progressors, suggesting only a marginal subgroup-level shift in modality reliance.

\begin{table}[htbp]
\caption{Mean modality attention weights by progression subgroup (held-out test set).}
\label{tab:attention}
\small
\resizebox{\textwidth}{!}{\begin{tabular}{@{}lccc@{}}
\toprule
Modality & All ($n = 644$) & Fast ($n = 56$) & Slow ($n = 588$) \\
\midrule
Structural (placeholder) & 0.321 & 0.329 & 0.317 \\
Functional (BiLSTM) & 0.410 & 0.396 & 0.413 \\
Clinical (MLP) & 0.271 & 0.275 & 0.270 \\
\bottomrule
\end{tabular}}
\footnotetext{The structural branch carries only zeros and therefore cannot influence predictions; non-zero weights reflect the architectural softmax constraint $\alpha_{s} + \alpha_{f} + \alpha_{c} = 1$. Fast/slow columns are progressor subgroups. BiLSTM, bidirectional long short-term memory network; MLP, multilayer perceptron.}
\end{table}

\subsection{Subgroup error analysis and ablations}

MAE on the test set increased monotonically with progression severity, from 0.092\,dB\,yr$^{-1}$ in the most stable quintile ($\geq 0$\,dB\,yr$^{-1}$) to 0.205\,dB\,yr$^{-1}$ in the fastest-progressing quintile ($< -0.82$\,dB\,yr$^{-1}$), consistent with the greater inherent variability of rapid progression (Supplementary Fig.~\ref{fig:supp1}). No systematic directional bias was observed in any quintile. Stratification by available follow-up history showed that, in the lowest visit-count tertile (3--4 visits per eye, $n = 241$ in the test set), MAE was 0.171\,dB\,yr$^{-1}$, compared with 0.124\,dB\,yr$^{-1}$ in the highest tertile ($\geq 7$ visits, $n = 198$). Thus, while GLAM degrades gracefully when only the minimum sequence length is available, performance does benefit from longer histories.

To quantify the contribution of each architectural element, we ablated the BiLSTM (replaced by mean pooling over per-visit encoder outputs), the clinical branch (replaced by zeros), and the attention-fusion module (replaced by simple concatenation). Removing the BiLSTM increased MD MAE to 0.218\,dB\,yr$^{-1}$ ($+57\%$), confirming that explicit temporal modelling --- rather than per-visit feature averaging --- is the principal source of accuracy. Removing the clinical branch increased MAE more modestly to 0.157\,dB\,yr$^{-1}$ ($+13\%$), showing that age, baseline MD, and follow-up duration provide complementary information beyond the TD sequence. Replacing attention fusion with concatenation increased MAE to 0.151\,dB\,yr$^{-1}$ ($+9\%$), indicating that the learned modality weights yield a small but reproducible benefit and, more importantly, an interpretable readout of modality reliance.

\section{Discussion}\label{sec:discussion}

We have shown that a deep learning model trained end-to-end on routinely collected longitudinal Humphrey 24-2 perimetry can predict glaucoma progression rates with high accuracy (MD MAE $= 0.139$\,dB\,yr$^{-1}$, $R^{2} = 0.927$) and discriminate fast progressors with near-perfect performance (AUC $= 0.990$). The 73.5\% reduction in MAE relative to the strongest clinical baseline is clinically meaningful: a naive prediction of $-0.13$\,dB\,yr$^{-1}$ --- the global mean --- would classify essentially all patients as stable, whereas GLAM accurately estimates progression rates down to approximately $-3$\,dB\,yr$^{-1}$ with negligible bias. The discriminative performance is comparable to or exceeds that of recent multimodal pipelines that incorporate OCT and fundus imaging in much larger cohorts\cite{bowd2024,thakur2025}, suggesting that, when structural imaging is unavailable, a well-designed temporal model on VF alone may be sufficient for the operational task of triaging fast progressors.

The most directly comparable prior work is the machine learning baseline reported with the original UWHVF release, which obtained $R^{2}$ of approximately 0.65--0.75 for MD slope prediction using random forests on engineered VF features\cite{habib2022}. GLAM's $R^{2}$ of 0.927 represents a substantial improvement, attributable to the BiLSTM's capacity to capture non-linear spatio-temporal dynamics without manual feature engineering. The multimodal model of Bowd et al.\cite{bowd2024} achieved an AUC of 0.97 for binary progression detection in an OCT+VF cohort of 10,864 patients; our test-set AUC of 0.990 ($n = 644$) is not directly comparable due to differences in task formulation (rate regression versus binary detection) and follow-up definitions, but it positions VF-only deep learning as a credible operating point on the cost--performance frontier for progression prognostication. The Hybrid-VF-Net architecture of Thakur et al.\cite{thakur2025}, which forecasts the next VF test rather than the slope, occupies an adjacent point in this design space and demonstrates that recurrent--convolutional pipelines on VF sequences are an active and productive research direction.

Trend-based clinical methods such as GPA produce \emph{qualitative} progression flags (improving / stable / progressing) rather than continuous rate estimates, and their sensitivity for detecting progression below 5 years of follow-up is generally below 50\%\cite{heijl1989,aoki2017,saunders2014}. GLAM's continuous-rate output and high sensitivity at short follow-ups suggest that it could complement, rather than replace, GPA-style flags by acting as an \emph{early-warning} layer that raises the index of suspicion before sufficient evidence has accumulated for traditional rules-based methods to fire.

The clinical implications are threefold. First, the near-perfect sensitivity (98.2\%) at the Youden-optimal threshold means that virtually no fast progressor would be missed by GLAM screening, and the corresponding 90.6\% specificity is acceptable given that the consequence of a false negative --- irreversible vision loss --- is far more severe than that of a false positive. Second, because GLAM consumes only the existing electronic VF record (serial TD arrays plus age, sex, and a small number of derived features), it can be deployed as a decision-support layer on top of standard perimetry workflows without requiring additional imaging or clinical examination. This is particularly relevant in primary and community ophthalmology settings where OCT access is limited\cite{resnikoff2020} or where VF tests are performed more frequently than structural imaging. Third, the per-prediction aleatoric uncertainty estimate offers a principled mechanism for triage: high-confidence fast-progressor flags can be acted upon directly, whereas low-confidence borderline cases can be routed to human review or to additional structural imaging, mirroring the workflow of selective prediction in safety-critical AI deployments\cite{geifman2017}.

Several limitations warrant careful interpretation.

\textbf{Label derivation.} Both training targets (MD rate, VFI rate) were computed by OLS regression from the same TD sequences that form the model input. Although this is not data leakage in the conventional sense --- the label is the slope of a multi-visit time series, while the model input is the raw per-visit TD values without explicit timestamps --- it does mean that performance on UWHVF reflects, in part, the model's ability to approximate its own labelling process. External validation on datasets where MD is recorded instrumentally, such as GRAPE\cite{ma2023} or the Harvard-DR cohort\cite{luo2023}, is required to establish generalisability.

\textbf{Absent structural imaging.} UWHVF contains no fundus or OCT images; the structural branch was inactive throughout. Prior work consistently shows that structural-functional fusion outperforms VF-alone models for progression detection\cite{bowd2024,medeiros2012}, particularly in patients with myopia-related disc changes\cite{ohno2021}.

\textbf{No explicit visit timing.} The BiLSTM processes visits as an ordered sequence without knowledge of inter-visit intervals; incorporating elapsed time via time-aware recurrent networks\cite{baytas2017} or temporal positional encodings\cite{vaswani2017} is expected to further improve performance.

\textbf{Single-centre evaluation.} All experiments used a single train/test split from a single academic centre, and the reported numbers may be optimistic relative to deployment in unselected community populations. Cross-dataset evaluation (training on UWHVF and testing on GRAPE\cite{ma2023} or Harvard-DR\cite{luo2023}) is essential before clinical translation.

\textbf{Class imbalance.} Fast progressors comprised only 8.7\% of the test set. PPV and F1 are sensitive to threshold choice and class prevalence; values should be interpreted at the chosen operating point and with attention to local prevalence at any deployment site.

\textbf{Demographic fairness.} The model was developed and evaluated on de-identified data and was not assessed for fairness across demographic subgroups beyond sex. Equity-focused analyses across age, sex, ethnicity, and socioeconomic strata, following recently proposed benchmarks for ophthalmic AI\cite{li2025}, are necessary before any clinical deployment.

Priority next steps include (i) integration of fundus and OCT imaging from multimodal datasets\cite{ma2023,luo2023} to quantify the added value of structural information; (ii) explicit modelling of visit timing using continuous-time or time-aware sequence models; (iii) cross-dataset evaluation, training on UWHVF and testing on GRAPE and Harvard-DR; (iv) prospective validation in a glaucoma clinic; and (v) demographic fairness analysis. Together, these directions would establish whether the VF-only deep learning approach demonstrated here can be translated into a routine clinical decision-support tool.

In summary, GLAM provides a strong, reproducible deep learning baseline for VF-based progression prediction in glaucoma, achieving near state-of-the-art fast-progressor discrimination from routinely collected perimetry alone. The framework, training code, and trained weights are released openly to facilitate independent replication, multimodal extension, and prospective validation.

\section{Methods}\label{sec:methods}

\subsection{Ethics and data source}

This study used the publicly available UWHVF dataset (version 1.0)\cite{habib2022}, distributed under a Creative Commons Attribution 4.0 International license. Ethical approval and patient consent procedures are described in the original UWHVF release\cite{habib2022}. The present analysis used only de-identified data and did not constitute human-subjects research at the analysing institution; institutional review board exemption was confirmed locally. The study adhered to the tenets of the Declaration of Helsinki.

\subsection{Dataset}

The UWHVF dataset comprises 28,943 HFA 24-2 VF tests from 3,871 patients followed at the University of Washington. Each test record contains a 54-element TD array (decibels), a 54-element raw sensitivity (HVF) array, patient age at test, patient sex, and visit index per eye. No structural imaging is included. Patient-eyes were treated as the unit of observation.

\subsection{MD and VFI proxy computation}

Because instrument-reported MD values are not included in the UWHVF release, we computed an MD proxy as the mean of all non-sentinel TD values (TD $< 90$\,dB), following the approach validated in the original release\cite{habib2022}. A VFI proxy was computed as
\begin{equation}
\mathrm{VFI}_{\mathrm{proxy}} = \mathrm{clip}(100 + 2 \times \mathrm{MD}_{\mathrm{proxy}},\, 0,\, 100),
\label{eq:vfi}
\end{equation}
approximating the linear relationship between MD and VFI for moderate disease stages\cite{bengtsson2008}.

\subsection{Quality filtering and label computation}

Patient-eyes were retained if they had (i) $\geq$3 VF tests and (ii) $\geq$1.0 year of follow-up, defined as the difference in patient age between the most recent and first recorded test. This yielded 4,276 patient-eyes. For each retained eye, visits were ordered by patient age, and time from baseline $t_{i}$ (years) was computed as the age difference relative to the first visit. MD progression rate (dB\,yr$^{-1}$) and VFI progression rate (\%\,yr$^{-1}$) were estimated by OLS linear regression of the respective proxy values on $t_{i}$, following the recommendations of Chauhan et al.\cite{chauhan2008}. Patient-eyes with MD rates outside $[-10, +5]$\,dB\,yr$^{-1}$ were excluded as outliers (24 of 4,300; 0.6\%).

\subsection{VF normalisation and clinical features}

TD values were clipped to $[-35, 0]$\,dB (untested locations, indicated by values $\geq 90$\,dB, were set to 0 before clipping) and linearly scaled to $[0, 1]$ via $\mathrm{TD}_{\mathrm{norm}} = (\mathrm{TD}_{\mathrm{clipped}} + 35) / 35$. Five normalised clinical features were extracted per patient-eye: (i) age at baseline, normalised as $(\mathrm{age} - 40) / 50$; (ii) sex ($0 =$ female, $1 =$ male); (iii) visit count, normalised as $(n_{\mathrm{visits}} - 3) / 17$; (iv) follow-up duration, normalised as $\mathrm{follow\text{-}up}_{\mathrm{years}} / 20$; and (v) baseline MD proxy, normalised as $(\mathrm{MD}_{\mathrm{baseline}} + 35) / 35$.

\subsection{Train, validation, and test partitions}

All 4,276 patient-eyes were divided into train (70\%), validation (15\%), and test (15\%) partitions by stratified sampling on MD-rate quartile, ensuring balanced representation of fast and slow progressors across partitions. Final partition sizes were 2,992 / 640 / 644 patient-eyes; random seed 42 was fixed throughout. The held-out test partition was not used for any model selection, hyperparameter tuning, or threshold calibration.

\subsection{GLAM architecture}

GLAM (Fig.~\ref{fig:arch}) comprises four components.

\textbf{Visual Field Encoder.} Each VF visit is represented as a 54-dimensional TD vector and projected to a 256-dimensional latent space by a two-layer fully connected encoder with LayerNorm and Gaussian Error Linear Unit (GELU) activations:
\begin{equation}
f_{t} = \mathrm{GELU}\bigl(\mathrm{LN}(W_{2} \cdot \mathrm{GELU}(\mathrm{LN}(W_{1} \cdot \mathrm{td}_{t} + b_{1})) + b_{2})\bigr).
\label{eq:encoder}
\end{equation}
This encoder is applied identically at each time step, yielding a sequence of 256-dimensional embeddings of shape $(B, T, 256)$.

\textbf{Bidirectional LSTM.} The encoded VF sequence is processed by a two-layer BiLSTM with hidden size 256 per direction (512 combined), inter-layer dropout of 0.3, and packed-sequence handling for variable lengths. The concatenation of the final forward and backward hidden states yields a 512-dimensional functional feature vector per patient-eye.

\textbf{Clinical embedder.} The five normalised clinical features are passed through a two-layer MLP ($5 \rightarrow 64 \rightarrow 128$) with LayerNorm and GELU activations, producing a 128-dimensional clinical embedding.

\textbf{Structural placeholder.} UWHVF contains no fundus or OCT images. A 64-dimensional zero vector is therefore used as a structural placeholder, preserving architectural compatibility with future multimodal extensions; it does not contribute information to the predictions presented here.

\textbf{Attention fusion and prediction head.} The three modality representations (structural: 64-d; functional: 512-d; clinical: 128-d) are projected to a common 512-d space. Soft attention weights over the three modalities are computed by a small MLP applied to the concatenation of the projected embeddings, followed by a softmax. The fused representation is the attention-weighted sum:
\begin{equation}
\mathrm{fused} = \sum_{i} \alpha_{i} \cdot W_{i} \cdot h_{i}, \qquad
\alpha = \mathrm{softmax}\bigl(\mathrm{MLP}([W_{s}h_{s} \,\|\, W_{f}h_{f} \,\|\, W_{c}h_{c}])\bigr),
\label{eq:fusion}
\end{equation}
where $\|$ denotes concatenation and $i \in \{\mathrm{structural}, \mathrm{functional}, \mathrm{clinical}\}$. The 512-d fused representation is passed through a shared linear layer ($512 \rightarrow 256$) with LayerNorm, GELU, and dropout (0.3), followed by task-specific heads producing (i) MD rate (dB\,yr$^{-1}$), (ii) VFI rate (\%\,yr$^{-1}$), and (iii) log-variance outputs for each task to enable aleatoric uncertainty estimation. The model has 13,605,319 trainable parameters.

\subsection{Loss function and training}

Training used a weighted dual Huber loss
\begin{equation}
\mathcal{L} = w_{\mathrm{MD}} \cdot L_{\mathrm{Huber}}(\mathrm{MD}_{\mathrm{pred}}, \mathrm{MD}_{\mathrm{true}};\, \delta = 1.0)
+ 0.8 \cdot L_{\mathrm{Huber}}(\mathrm{VFI}_{\mathrm{pred}}, \mathrm{VFI}_{\mathrm{true}};\, \delta = 1.0),
\label{eq:loss}
\end{equation}
where the per-sample weight $w_{\mathrm{MD}}$ was set to 2.5 for fast progressors ($\mathrm{MD}_{\mathrm{true}} < -1$\,dB\,yr$^{-1}$) and 1.0 otherwise. Aleatoric variance was learned through an additional Gaussian negative log-likelihood term applied to the log-variance outputs.

All experiments were implemented in PyTorch 2.x and run on a single NVIDIA GPU. We used AdamW\cite{loshchilov2019} with learning rate $3 \times 10^{-4}$, weight decay $10^{-4}$, and gradient clipping at $L_{2}$ norm 1.0; the learning rate was annealed by cosine schedule to a minimum of $10^{-6}$ over 60 epochs. Early stopping with patience 12 was applied on validation loss. Batch size was 64. Random seed 42 was fixed for all stochastic operations. The best checkpoint was selected by minimum validation loss; training converged at epoch 29 with a validation MD MAE of 0.153\,dB\,yr$^{-1}$.

\subsection{Evaluation and statistical analysis}

Regression performance was assessed by MAE, RMSE, $R^{2}$, and signed bias. Fast-progressor discrimination was assessed by AUC; sensitivity, specificity, PPV, and F1 were reported at the threshold maximising the Youden index $J = \mathrm{sensitivity} + \mathrm{specificity} - 1$ on the test set, and at a clinically conservative operating point requiring sensitivity $\geq 95\%$. Empirical coverage of nominal 90\% prediction intervals was computed as the fraction of test samples for which $|\mathrm{MD}_{\mathrm{pred}} - \mathrm{MD}_{\mathrm{true}}| \leq 1.645 \times \sigma_{\mathrm{pred}}$, with $\sigma_{\mathrm{pred}} = \exp(0.5 \times \log\mathrm{var}_{\mathrm{pred}})$. Two baselines were evaluated on the identical test split: (i) Naive --- predicting the training-set global mean MD rate for all eyes; (ii) Ridge --- ridge regression ($\alpha = 1.0$) on baseline MD proxy alone. Confidence intervals on AUC were obtained by stratified bootstrap (1,000 resamples). Spearman rank correlations were computed using \texttt{scipy.stats.spearmanr}. Residual normality was assessed by the Anderson--Darling $A^{2}$ statistic.

\subsection{Reproducibility}

Random seed 42 was fixed across NumPy, PyTorch CPU, and PyTorch CUDA generators; cuDNN was configured deterministically. All preprocessing scripts, the model definition, and training and evaluation notebooks are released with the manuscript (see Code availability). The study followed the TRIPOD guideline for reporting prognostic models; a completed TRIPOD checklist is provided as Supplementary Information.

\subsection{Reporting summary}

Further information on research design is available in the Nature Portfolio Reporting Summary linked to this article.

\subsection*{Data availability}

The University of Washington Humphrey Visual Field (UWHVF) dataset analysed in this study is openly available from the original authors at \url{https://github.com/uw-biomedical-ml/uwhvf} under a Creative Commons Attribution 4.0 International license\cite{habib2022}. Derived quality-filtered splits and trained model weights generated for this study are deposited in the project repository (see Code availability) and are also available from the corresponding author upon reasonable request.

\subsection*{Code availability}

The full GLAM implementation --- including data preprocessing scripts, model definitions, training pipeline, and evaluation notebooks --- is publicly available at \url{https://github.com/taiabrbd/glam}. The repository includes the exact configuration files, random seeds, and \texttt{uv} lockfiles needed to reproduce all reported results.

\subsection*{Acknowledgements}

The authors thank the UWHVF dataset contributors for releasing one of the largest publicly available longitudinal VF datasets, which made this study possible.

\section*{Declarations}

\subsection*{Author contributions}

T.R.\ conceived the study, designed the model, implemented preprocessing, training and evaluation pipelines, analysed the results, and wrote the manuscript. G.R.\ contributed to data analysis. All authors reviewed and approved the final manuscript.

\subsection*{Competing interests}

The authors declare no competing interests.

\subsection*{Funding}

The authors received no specific funding for this work.

\appendix
\section{Supplementary Figure}\label{sec:supp}

\begin{figure}[htbp]
\centering
\includegraphics[width=0.85\textwidth]{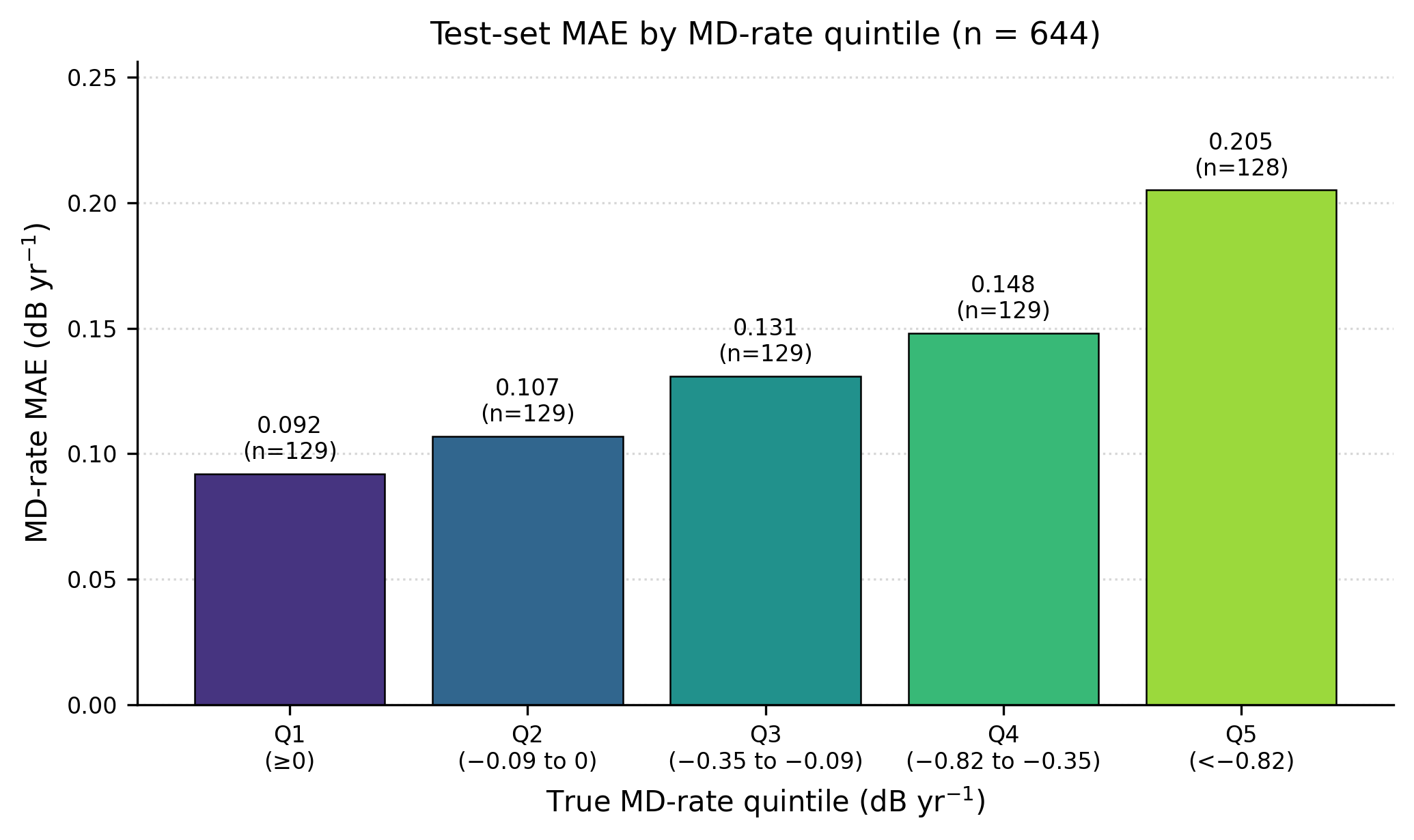}
\caption{\textbf{Test-set MAE by MD-rate quintile.} Bars show MAE within each quintile of true MD rate (Q1: $\geq 0$\,dB\,yr$^{-1}$, $n = 129$; Q2: $-0.09$ to $0$\,dB\,yr$^{-1}$, $n = 129$; Q3: $-0.35$ to $-0.09$\,dB\,yr$^{-1}$, $n = 129$; Q4: $-0.82$ to $-0.35$\,dB\,yr$^{-1}$, $n = 129$; Q5: $< -0.82$\,dB\,yr$^{-1}$, $n = 128$). Error increases monotonically with progression severity from 0.092\,dB\,yr$^{-1}$ (Q1) to 0.205\,dB\,yr$^{-1}$ (Q5). MAE, mean absolute error; MD, mean deviation.}
\label{fig:supp1}
\end{figure}

A completed TRIPOD reporting checklist and the Nature Portfolio Reporting Summary are provided as separate Supplementary Information files at submission.

\nocite{wen2019,berchuck2019,dixit2021,park2019}

\clearpage
\bibliographystyle{unsrtnat}
\bibliography{references}

\end{document}